\documentclass[letterpaper, 10 pt, journal, twoside]{ieeetran}

\pdfoutput=1

\usepackage{hyperref}
\usepackage{amsmath}
\usepackage{amssymb}
\usepackage{graphicx}
\usepackage{fixltx2e}
\usepackage{booktabs}
\usepackage{subcaption}
\usepackage[export]{adjustbox}
\usepackage{siunitx }
\usepackage{tikz}

\graphicspath {{figures/}}

\newcommand{\etal}{et al.~}
\newcommand{\likelihood}{$p_{\mathbf{\theta}}(\mathbf{x} | \mathbf{z})$}

\DeclareMathOperator*{\argmax}{arg\,max}
\DeclareMathOperator{\E}{\mathbb{E}}

\IEEEoverridecommandlockouts
\title{\LARGE \bf
Modeling Grasp Motor Imagery through \\Deep Conditional Generative Models }

\author{Matthew Veres, Medhat Moussa, and Graham W.~Taylor
\thanks{
Manuscript received: September, 10, 2016; Revised November, 30, 2016; Accepted December, 27, 2016.}%
\thanks{This paper was recommended for publication by Editor Han Ding upon evaluation of the Associate
Editor and Reviewers' comments. This work is supported by the Natural Sciences and Engineering 
Research Council of Canada, and the Canada Foundation for Innovation.}
\thanks{
Authors are with the School of Engineering,
        University of Guelph, 50 Stone Road East. Guelph, Ontario, Canada.
        {\tt\small \{mveres,mmoussa,gwtaylor\}@uoguelph.ca}}%
\thanks{Digital Object Identifier (DOI): see top of this page.}%
}

\newcommand\copyrighttext{%
  \footnotesize \textcopyright 
2017 IEEE. Personal use of this material is permitted. Permission from IEEE must be
obtained for all other uses, in any current or future media, including
reprinting/republishing this material for advertising or promotional purposes, creating new
collective works, for resale or redistribution to servers or lists, or reuse of any copyrighted
component of this work in other works.
}
\newcommand\copyrightnotice{%
\begin{tikzpicture}[remember picture,overlay]
\node[anchor=south,yshift=5pt] at (current page.south) {\fbox{\parbox{\dimexpr\textwidth-\fboxsep-\fboxrule\relax}{\copyrighttext}}};
\end{tikzpicture}%
}

\begin{document}

\maketitle
\copyrightnotice

\begin{abstract}
  Grasping is a complex process involving knowledge of the object, the
  surroundings, and of oneself. While humans are able to integrate and process
  all of the sensory information required for performing this task, equipping
  machines with this capability is an extremely challenging endeavor. In this
  paper, we investigate how deep learning techniques can allow us to translate
  high-level concepts such as \textit{motor imagery} to the problem of robotic
  grasp synthesis. We explore a paradigm based on generative models for learning
  integrated object-action representations, and demonstrate its capacity for
  capturing and generating multimodal, multi-finger grasp configurations on a
  simulated grasping dataset.
\end{abstract}

\IEEEpeerreviewmaketitle


\section{Introduction}

\IEEEPARstart{H}{umans} have an innate ability for performing complex grasping maneuvers. Often
times, these maneuvers are performed unconsciously, where object dynamics are
unknown or continuously changing through time. This ability also manifests where
objects themselves may be either similar or novel to those previously
encountered. Given some prior experience on grasping an object, it seems highly
unlikely that humans learn from scratch how to grasp each new object that is
presented to them. Rather, we posit that this ability is driven through both
motor and object representations, allowing for abstract generalizations and
efficient transference of skills among objects.

In robotics, grasping and manipulation is a critical and challenging problem.
Its difficulty stems from variability in an object's shape and
physical properties, the gripper capabilities, and task requirements. As such,
most industrial applications require robots to use myriad gripping fixtures or
end-of-arm tools to grasp various objects. But as robots expand to applications
in non-structured environments (both in industry and beyond), advanced grasping
skills are needed.

Currently there are several difficulties in actually \textit{learning} how to
grasp. First, the problem is fundamentally a many-to-many mapping. An object can
be grasped in many ways that are all equivalent, while the same grasp
configuration can be used to grasp many objects. There is a need to maintain
this many-to-many mapping to enable the robot to grasp objects under uncertainty
and in highly cluttered environments. Second, while the object shape and
location can be obtained from perception, grasping is often challenged by inherent
characteristics of the object such as surface friction, weight, center of mass,
and finally the actual functionality of the object. All of these factors are
only known \textit{after} the object is touched and the grasp is started.

In this paper, we propose to learn a new concept that we refer to as the
\textit{grasp motor image} (GMI). The GMI combines object perception and a
learned prior over grasp configurations, to synthesize new grasps to apply to a
different object. We liken this approach to the use of motor representations
within humans.  Specifically, we focus on the use of motor imagery for creating
internal representations of an action, which requires some knowledge or
intuition of how an object may react in a given scenario.

We show that using recent advances in deep learning (DL), specifically deep
conditional generative models \cite{sohn2015learning} and the Stochastic
Gradient Variational Bayes (SGVB) framework \cite{kingma2013auto}, we can
capture multimodal distributions over the space of grasps conditioned on visual
perception, synthesizing grasp configurations with minimal additional complexity
compared to conventional techniques such as convolutional neural networks
(CNNs). We quantitatively compare our approach to the discriminative CNN
baseline and other generative models and qualitatively inspect samples generated
from the learned distribution.

\subsection{Contributions}

Most work within deep learning and robotic grasp synthesis has focused in one
form or another on the \textit{prediction} of grasp configurations given visual
information. The goal of this work is to show how having an idea of the
properties characterizing an object, and an idea of how a similar object was
grasped previously, a unified space can be constructed that allows grasps
to be generated for novel objects.

A second contribution of this work is a \textit{probabilistic} framework,
leveraging deep architectures to learn multimodal grasping distributions for
multi-fingered robotic hands. Grasping is inherently a many-to-many mapping, yet
as we show in this paper, na\"{\i}vely applying mainstream deep learning
approaches (e.g.~convolutional neural networks) may fail to capture these
complex distributions or learn in these scenarios without some stochastic
components. Here, we demonstrate the feasibility of deep generative models for
capturing multimodal distributions conditional on visual input, and open the
door to future work such as the integration of other sensory modalities.


\section{Background and Related work}

Within robotics, our work shares some similarities to the notion of
\textit{experience databases} (ED), where prior knowledge is used to either
synthesize grasps partially (i.e.~through constraints or algorithmic priors
\cite{mahlerdex}) or fully, by using previously executed grasps. Our work is
also similar in spirit to the concept of \textit{Grasp Moduli Spaces}
\cite{pokorny2013grasp}, which define a continuous space for grasp and shape
configurations.

Yet, instead of classical approaches with EDs (which require manual
specification of storage and retrieval methods), our approach allows these
methods to be learned automatically, for highly abstract
representations. Further, our approach to constructing this ``grasp space'' is
to collect realistic data on object and grasp attributes using sensory
modalities and hand configurations, and learn the encoding space as an
integrated object-action representation.

\subsection{Motor imagery}

Motor imagery (MI) and motor execution (ME) are two different forms of motor
representation. While motor execution is an external representation (physical
performance of an action), motor imagery is the use of an internal
representation for mentally \textit{simulating} an action
\cite{jeannerod2001neural,mulder2007motor}.

As MI is an internal process, assessing MI performance is typically done by
analyzing behavioural patterns (such as response time for a given task), or by
visualizing neural activity using techniques such as functional magnetic
resonance imagining (fMRI). Using these techniques, many studies have reported
overlapping activations in neural regions between both MI and ME
(e.g.~\cite{hanakawa2008motor,sharma2013does}, review:
\cite{munzert2009cognitive}), suggesting that at some level, participating in
either MI or ME affects some amount of shared representation. These findings
lead credence to the hypothesis that, for example, mentally simulating a
grasping task shares some measure of similar representation to actually
performing the grasp itself.

Among many studies that have examined this phenomenon, we highlight one by Frak
\etal \cite{frak2001orientation} who explored it in the context of which frame
of reference is adopted during implicit (unconcious) MI performance. The authors
presented evidence that even though MI is an internal process, participants
mentally simulating a grasp on a water container did so under real-world
biomechanical constraints. That is, grasps or actions that would have been
uncomfortable to perform in the real world (e.g.~due to awkward joint
positioning) were correlated with responses of the mentally simulated action.

\subsection{Learning joint embeddings and object-action interactions}

Learning joint embeddings of sensory and motor data is not new. It has
been proposed, e.g., by Uno \etal \cite{uno1993integration}, who
attempted to learn a joint embedding between visual and motor information. In
the field of robotics, other works that have used multimodal embeddings include
Sergeant \etal \cite{sergeantmultimodal} for control of a mobile robot, Noda
\etal \cite{noda2014multimodal} for behaviour modeling in a humanoid robot, and
recently Sung, Lenz, and Saxena \cite{sung2015deep} who focus on transferring
trajectories.

Congruent to learning joint embeddings, there has also been work in robotic
grasping on learning how objects and actions interact with each other, and what
effects they cause. Bayesian networks have been used to explore these effects by
Montesano \etal \cite{montesano2008learning}, who use discretized high-level
features and learn the structure of the network as part of a more generalized
architecture. Other work with Bayesian networks include Song \etal
\cite{song2011multivariate}, who explore how large input spaces can be
discretized efficiently for use in grasping tasks.

Our approach differs significantly from these works in the scale of data,
application, and network structures. Further, with respect to Bayesian networks,
we work with continuous data without any need for discretization.

\subsection{Deep learning and robotic grasping}

The majority of work in DL and robotic grasping has focused on the use of
parallel-plate effectors with few degrees of freedom. Both Lenz
\cite{lenz2015deep}, and Pinto \cite{pinto2015supersizing} formulate grasping as
a \textit{detection} problem, and train classifiers to predict the most likely
grasps through supervised learning. By posing grasping as a detection problem,
different areas of the image could correspond to many different grasps and fit
with the multimodality of grasping; yet, to obtain multiple grasps for the same
image patch, some form of stochastic components or \emph{a priori}
knowledge of the types of grasps is required.

Mahler \etal \cite{mahlerdex} approach the problem of grasping through the use
of deep, multi-view CNNs to index prior knowledge of
grasping an object from an experience database.  Levine \etal
\cite{levine2016learning} work towards the full motion of grasping by linking
the prediction of motor commands for moving a robotic arm with the probability
that a grasp at a given pose will succeed. Other work on full-motion robotic
grasping includes Levine \cite{levine2015end} and Finn \cite{finn2015learning}
who learn visuomotor skills using deep reinforcement learning.

DL and robotic grasping have also recently extended to the domain of
multi-fingered hands. Kappler \etal \cite{kappler2015leveraging} used
DL to train a classifier to predict grasp stability of the Barrett hand under
different quality metrics. In this work, rather than treating grasping as a
classification problem, we propose to predict \textit{where} to grasp an
object with a multi-fingered hand, through a gripper-agnostic representation of
available contact positions and contact normals.

\section{Grasp motor imagery}

Although the field is far from a consensus on the best uses and structures for
DL and robotic grasping, most work appears to use deep architectures
for processing visual information. At the core of our approach is the
autoencoder structure. We briefly review the principles of this method before
reviewing the probabilistic models considered herein.

\subsection{Autoencoders}

\begin{figure}[htb!]
  \centering
	\includegraphics[width=0.3\textwidth]{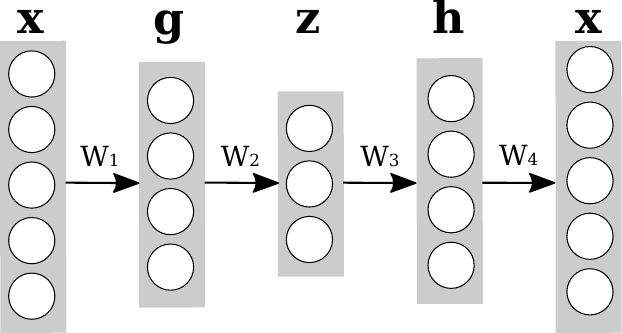}
    \caption{Autoencoder structure}
  	\label{fig:ae}
\end{figure}

An autoencoder (AE) is an unsupervised deep learning algorithm that attempts to
dissociate latent factors in data using a combination of encoding and decoding
networks (Figure \ref{fig:ae}). In the encoding stage, an input $\mathbf{x}$ is
mapped through a series of (typically) constricting nonlinear hidden layers to
some low-dimensional latent representation of the input $f(\mathbf{x})$. In
autoencoders that have an odd number of layers, this layer is often denoted by
$\mathbf{z}$. The decoding phase forces the learned representation to be
meaningful by mapping it to some reconstruction $g(f(\mathbf{x}))$ of the
original input. Training proceeds by iteratively encoding and decoding a
datapoint, measuring the difference between the reconstruction and original
input, and backpropagating the error to update the weights of both the encoder
and decoder.

While the AE is capable of untangling the latent factors of complex data one
limitation of this architecture is that autoencoders are inherently
\textit{deterministic} in nature; a single input will always produce the exact
same latent representation and reconstruction. For tasks such as robotic
grasping, which is inherently a many-to-many mapping (i.e.~many different
representations may map to many different grasps), deterministic networks such
as the AE are unsuitable for implementing such a mapping in a generative sense.

\subsection{Variational autoencoders}

\begin{figure}[htb!]
    \centering
    \begin{subfigure}{0.17\textwidth}
        \centering
        \includegraphics[width=\textwidth]{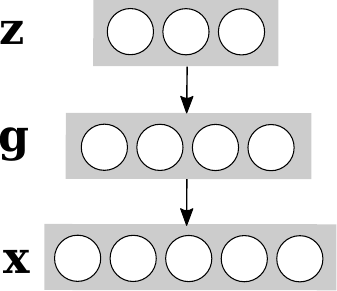}
  	    \caption{Generator network}
        \label{fig:vae_gen}
    \end{subfigure}
    \hspace{0.2cm}
    \begin{subfigure}{0.175\textwidth}
        \centering
        \includegraphics[width=\textwidth]{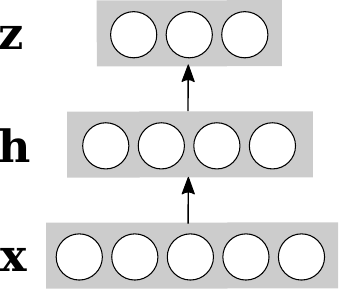}
  	    \caption{Recognition network}
        \label{fig:vae_recog}
    \end{subfigure}
\caption{Variational AE generator and recognition networks.}
\label{fig:variational_autoencoder}
\end{figure}

The variational autoencoder (VAE) \cite{kingma2013auto,rezende2014stochastic} is
a directed graphical model composed of \textit{generator} and
\textit{recognition} networks (Figure \ref{fig:variational_autoencoder}). The
goal of the recognition network is to learn an approximation to the intractable
posterior $p_{\theta}(\mathbf{z|x})$ by using an approximate inference network
$q_{\phi}(\mathbf{z|x})$ through some nonlinear mapping, typically parametrized
as a feedforward neural network. The generator network takes an estimate of
$\mathbf{z}$, and learns how to generate samples from
$p_{\theta}(\mathbf{x|z})$, such that
$p_{\theta}(\mathbf{x})=\sum_{\mathbf{z}}p(\mathbf{x}|\mathbf{z})p(\mathbf{z})$
approximates the data distribution $p(\mathbf{x})$.

These networks can be composed in a fashion similar to the classical AE, where
the recognition network forms a kind of encoder, and generator network
constitutes the decoder. This is, in fact the proposed method of training them
within the SGVB framework \cite{kingma2013auto}, which adds only the complexity
of sampling to classical methods, and still performs updates using the
backpropagation algorithm.

The objective of the VAE shares some similarities to the classical AE; that is,
in the case of continuous data the optimization of a squared error objective
along with an additional regularizing term. This objective is denoted
as the \textit{variational lower bound} on the marginal likelihood:
\begin{align}
\label{eqn:vae_lb}
\log p_{\theta}(\mathbf{x}) & \geq  -D_{KL}(q_{\phi}(\mathbf{z|x})||p_{\theta}(\mathbf{z})) \\+
& \E_{q_{\phi}(\mathbf{z|x})} \Big[ \log p_{\theta}
  (\mathbf{x|z})\Big] \nonumber
\end{align}

\noindent where $D_{KL}$ is the KL-Divergence between the encoding
and prior distribution, analagous to a regularization term in a standard
autoencoder.
The form of the \likelihood~ will depend on the nature of the data, but
typically is Bernoulli or Gaussian. Note that the VAE formulation only permits
continuous latent variables.

Commonly, both the prior $p_{\theta} (\mathbf{z})$ and encoding distributions
are chosen to be multivariate Gaussians with diagonal covariance matrices
$\mathcal{N}(\mathbf{\mu, \sigma^2I})$, in which case the recognition network
learns to encode $\mathbf{\mu}$ and $\mathbf{\sigma}$ for each latent variable.
This simple parameterization allows for the KL-divergence to be computed
analytically without the need for sampling. In order to compute the expectation,
the \textit{reparameterization trick} introduced by Kingma and Welling
\cite{kingma2013auto} reparameterizes (a non-differentiable) $\mathbf{z}$
through some differentiable function $g_{\phi}(\mathbf{x,\epsilon})$:
\begin{equation}
    \label{eqn:reparam}
    \mathbf{z} = \mathbf{\mu} + \mathbf{\sigma}\mathbf{\epsilon}
\end{equation}

\noindent where $\epsilon$ is sampled from the noise distribution $p(\epsilon) =
\mathcal{N}(0,\mathbf{I})$, and $\mathbf{\mu}$, $\mathbf{\sigma}$
are the mean and standard deviation of the encoding distribution respectively.
Thus, an estimate of the lower bound can be computed according to:
\begin{align}
\label{eqn:vae_lb_approx}
	\mathcal{L}_{\text{VAE}}(\mathbf{\theta, \phi; x}) & =
	-D_{KL}(q_{\phi}(\mathbf{z|x})||p_{\theta}(\mathbf{z})) \\ +
	& \frac{1}{L} \sum_{l=1}^{L}\log p_{\theta}(\mathbf{x|z^{(l)}}) \nonumber
\end{align}

\noindent where $\mathbf{z}$ is reparameterized according to Equation
\ref{eqn:reparam}, and $L$ is the number of samples drawn from the prior
distribution for computing the expectation\footnote{In many cases (such as large
minibatch sizes), only a single sample needs to be drawn.}.

Somewhat abstracted from view, but fundamental to the decision to pursue them in
this work is that VAEs are not only probabilistic models, but that the
stochasticity of $\mathbf{z}$ allows for modeling \textit{multiple modes} within
the data distribution. That is, sampling different $\mathbf{z}$'s may localize the network's
predictions to different high-probability regions of the reconstructed output
space. In our setup, we only condition on visual input and sample grasps, not
vice-versa. Therefore we do not explicitly treat the many-to-many
mapping. However, VAEs may also permit the joint modeling of grasp, vision, and
other perceptual inputs, such as tactile sensors. This is out of application
scope and reserved for future work.

\begin{figure*}[htb!]
\begin{subfigure}{0.99\textwidth}
  \includegraphics[width=\textwidth, center]{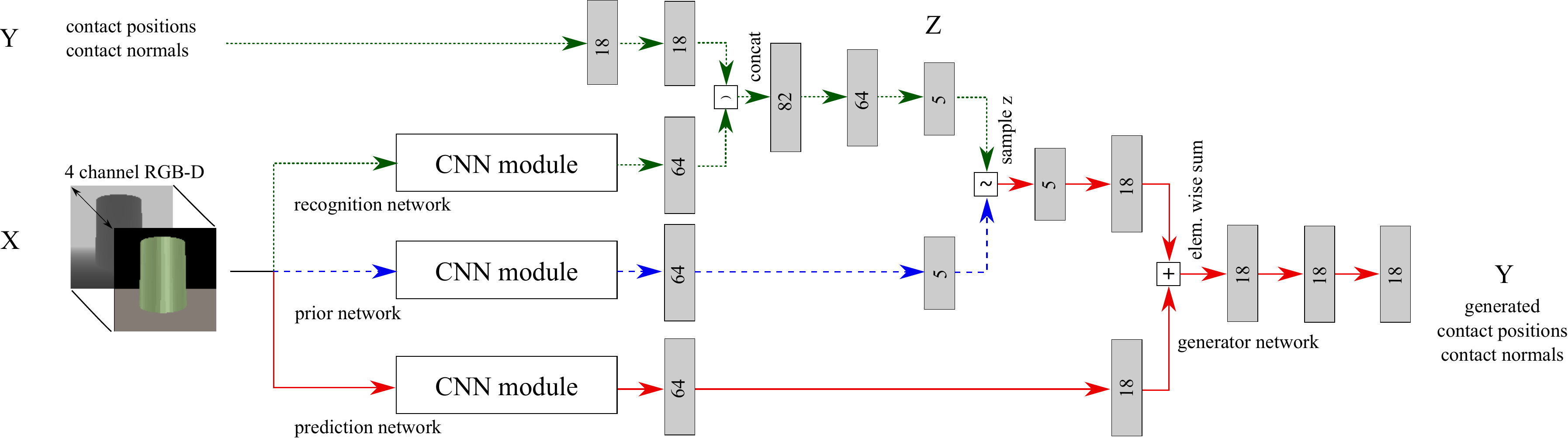}
  	\caption{CVAE architecture we use in our experiments.
  	Dotted arrows denote components
  			used during training, dashed arrows are components used during testing,
  			and solid arrows are used for both training and
                        testing. The CNN Module is expanded in (\subref{fig:prior_network}).}
   \label{fig:generator_network}
\end{subfigure}
\begin{subfigure}{0.99\textwidth}
	\includegraphics[width=\textwidth,center]{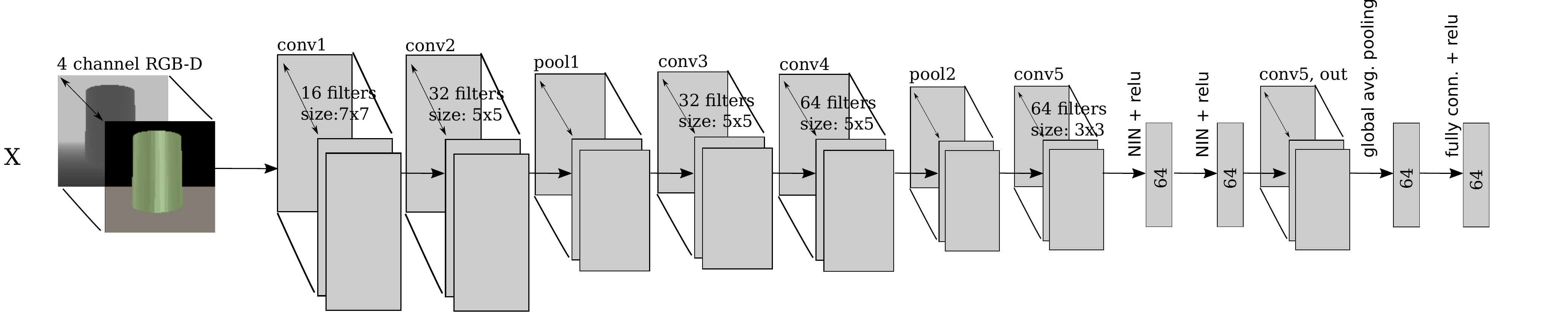}
    \caption{CNN module for processing visual information with
             Network-in-Network (NIN) layers \cite{lin2013network}.}
  	\label{fig:prior_network}
\end{subfigure}%
\vspace{0.5cm}
\caption{Schematic of our method. (\subref{fig:generator_network}) During training, the recognition
         network $p(\mathbf{z|x,y})$ learns an encoding distribution for the
         image and contact positions/normals, while the generator network
         p($\mathbf{y|x,z}$) takes a sample z, along with a representation from
         the prediction network to generate a sample grasp. The prior network
         p($\mathbf{z|x}$) is used to regularize the distribution learned by the
         recognition network (via the KL-Divergence term of Equation \ref{eqn:cvae_lb_approx}), and is also used to sample a $\mathbf{z}$ during
         testing, as the network does not have access to grasp information. In
         the R-CVAE network, the structure of the prior network matches that of
         the recognition network, but takes a predicted grasp (made by the
         prediction network) as input. (\subref{fig:prior_network}) All visual
         recognition modules are CNNs with NIN.}
\label{fig:networks}
\end{figure*}
%

\subsection{Conditional variational autoencoders}
\label{sec:cvae}

The conditional variational autoencoder (CVAE) \cite{sohn2015learning},
is an extension of the VAE architecture to settings with more than one source of
information. Given an input $\mathbf{x}$, output $\mathbf{y}$, and latent
variables $\mathbf{z}$, conditional distributions are used for the recognition
network $q_{\phi}(\mathbf{z|x, y})$, generator network
$p_{\theta}(\mathbf{y|x,z})$, and for a prior network
$p_{\theta}(\mathbf{z|x})$\footnote{The prior network is a technique for
modulating the latent variables via some input data, and is used in place of the prior specified for the VAE.}. In this work,
$\mathbf{x}$ represents input visual information (i.e.~images), and $\mathbf{y}$
is the output grasp configuration as shown  in Figure \ref{fig:networks}.

The lower bound of the CVAE is similar to the VAE, except for the
addition of a second source of given information:
\begin{align}
\label{eqn:cvae_lb_approx}
    \mathcal{L}_{\text{CVAE}}(\mathbf{\theta, \phi; x, y}) &=
    -D_{KL}(q_{\phi}(\mathbf{z|x,y})||p_{\theta}(\mathbf{z|x})) \\ +
    & \frac{1}{L} \sum_{l=1}^{L}\log p_{\theta}(\mathbf{y|x,z^{(l)}}) .\ \nonumber
\end{align}

The structure of each of these networks (recognition, generator, and prior) is a
design choice, but following \cite{sohn2015learning}, we design both our
recognition and generator network to have a convolutional (CNN-based) pathway
from the image for its numerous benefits, among them, reducing the number of
free parameters. The typical way to evaluate such models is achieved through
estimates of the conditional log-likelihood (CLL), using either Monte Carlo
(Equation \ref{eqn:montecarlo_cll}) or importance sampling (Equation
\ref{eqn:importance_sampling_cll}), the latter typically requiring a fewer
number of samples.
\begin{align}
    \label{eqn:montecarlo_cll}
    p_{\theta}(\mathbf{y|x}) & \approx \frac{1}{S} \sum_{s=1}^Sp_{\theta}(\mathbf{y|x,z}^{(s)}), \\
    &\mathbf{z}^{(s)}\sim p_{\theta}(\mathbf{z|x}) \nonumber \\ \nonumber \\
     \label{eqn:importance_sampling_cll}
    p_{\theta}(\mathbf{y|x}) & \approx \frac{1}{S} \sum_{s=1}^S
    \frac{p_{\theta}(\mathbf{y|x,z^{(s)}})p_{\theta}(\mathbf{z}^{(s)} | \mathbf{x})}
    {q_{\phi}(\mathbf{z}^{(s)} | \mathbf{x,y})},  \\ & \mathbf{z}^{(s)}
\sim q_{\phi}(\mathbf{z |x,y}) \nonumber
\end{align}

\subsection{Grasp motor image}

Consider, for example, transferring prior motor experience to novel objects,
where the selection of an initial grasp was influenced by internal and external
perceptual object properties. One approach to transferring this knowledge could
be to directly use past experience -- existing in some high-dimensional space -- to
initialize priors within a system. A different approach, e.g.~in MI, could be to
access some latent, low-dimensional representation of past actions which are
shared among a variety of different situations.

Instead of learning a direct, image-\textit{to}-grasp mapping through neural
networks, we instead learn an (image-\textit{and}-grasp)-to-grasp mapping. Our
approach is intuitive: based on perceptual information about an object, and an
idea of how an object was previously grasped, we index a shared structure of
object-grasp pairs to synthesize new grasps to apply to an object. As shown in
Section \ref{sec:cvae} this scheme exists as a single module and is trainable in
a fully end-to-end manner. Further, the structure of the CVAE model allows us to
model and generate grasps belonging to not only one, but possibly many different
modes.

Building the grasp motor image only requires that object properties are captured
in some meaningful way. We use a single data-modality (i.e.~visual information)
which exploits CNN modules for efficient and effective visual representation.
Our continued interest lies in investigating how the GMI can be gradually
learned through multiple modalities; specifically, those that capture internal
properties and require object interaction (such as force and tactile data).

\section{Experimental Setup}

There are a few different directions that reasoning about grasp synthesis using
GMI affords. Due to abstractions at both the object and action level, we
hypothesize that the model should require fewer number of samples to run, and
evaluate this by restricting the amount of training data available to the model.
We also evaluate synthesized grasps on objects similar to those seen at
training time, along with families of objects the model has never seen before.

\subsection{Dataset}

We collected a dataset of successful, cylindrical precision robotic grasps using
the V-REP simulator \cite{VREP}, and object files provided by Kleinhans \etal
\cite{kleinhans2016G3DB} on a simulated ``picking'' task. While there are
several three-fingered grippers being employed for commercial and research
applications (e.g.~the ReFlex hand \url{http://labs.righthandrobotics.com}), a
number of recent works have adopted the Barrett hand
\cite{kappler2015leveraging,kleinhans2016G3DB}\footnote{Note that the Barrett
  hand is an underactuated, three-fingered gripper parameterized through a total
  of 8 joints and 4 degrees of freedom.}. In this work, we favour this gripper
for its ability to capture a variety of different grasps dependent on the shape
of an object.

The object files comprise a set of various everyday household items, ranging in
size from small objects such as tongs, to larger objects such as towels and
vases. Each of these object classes has a unique ancestral template, and
perturbations of these templates were performed to yield a number of meshes with
different shape characteristics. During data collection we assumed that all
objects are non-deformable, share the same friction value, and share the same
mass of \SI{1}{\kilogram}. In a simulated setting, these assumptions allow us to
direct our attention towards the effects of visual information on the GMI and
forgo properties that are better captured through e.g.~tactile sensory systems.

From each object class, we remove 1 object and place it into a test set. The
training set is comprised of 161 objects from the 20 most populated classes,
containing around 47,000 successful image/grasp pairs. From this set, we
randomly select 10\% to be used for validation during training. In order to
study the effects of generating grasps for novel objects, we partition this test
set into two distinct groups: objects that are \textit{similar} to those of the
training set (i.e.~belong to the same \emph{class} of object but are not the
same instance), and the other set comprised of object classes never encountered
during training (\textit{different}). The final dataset statistics are reported
in Table \ref{tab:dataset_info}.

\begin{table}[htbp]
  \centering
  \caption{Overview of dataset statistics}
    \begin{tabular}{rrr}
    \toprule
          & \# Objects & \# Instances \\
    \midrule
    Training files & 161   & 42,351 \\
    Testing files - Similar & 20    & 4,848 \\
    Testing files - Different & 53    & 8,851 \\
    \bottomrule
    \end{tabular}%
  \label{tab:dataset_info}%
\end{table}%

\subsubsection{Visual information}

Recorded for each successful grasp trial are RGB and Depth images (size
64$\times$64 pixels), as well as a Binary segmentation mask of equal size,
indicating the object's spatial location in the image. Each image collected uses
a simulated Kinect camera with primary axis pointing towards the object (along
the negative z-direction of the manipulator's palm), and the y-axis pointing
upwards. This configuration means that the same image could correspond to a
number of different grasps, and allows us to capture multimodality that may
exist within the grasp space.

\subsubsection{Definition of grasps}

Our experiments leveraged the three-fingered Barrett hand, and defines a grasp
as the 18-dimensional vector
$[\vec{p_1}, \vec{p_2}, \vec{p_3}, \vec{n_1}, \vec{n_2}, \vec{n_3}]$, where the
subscript denotes the finger, and the contact positions $p$ and normals $n$ each
have $(x, y, z)$ Cartesian components. While the contact positions specify where
to place the fingertips of a gripper, the purpose of the contact normals is to
describe the relative orientation. Note that with this parameterization, the
unique characteristics of a manipulator (i.e.~number of joints or degrees of
freedom) have been abstracted into the number of available fingertips and
renders the representation as being gripper-agnostic.

We encode all grasps into the object's reference frame \{O\}, which is obtained
by applying PCA on the binary segmentation mask. We assign the directional
vectors $\vec{O_z}$ and $\vec{O_y}$ coincident with the first and second
principal components respectively, and ensure that $\vec{O_x}$ always points
into the object. We define the object's centroid as being the mean $x_p$ and
$y_p$ pixel coordinates of the object.

\subsection{Learning}

We build our networks within the Lasagne framework
\cite{sander_dieleman_2015_27878}, using the Adam optimizer, a learning rate of
0.001, and a minibatch size of 100. Due to class imbalance in the dataset
(complicated by some objects being easier to grasp than others), we train with
class-balanced minibatches. We standardize all data to have zero mean and
unit-variance.

For the recognition network we use a 5-layer CNN, applying max pooling every 2
layers and using filter sizes of [7, 5, 5, 5, 3] and number of filters [16, 32,
32, 64, 64] respectively. The output of the convolution operations feeds into a
network-in-network structure with average pooling \cite{lin2013network} for
reducing dimensionality of the system, and is subsequently followed by a
64-neuron hidden layer. The outputs from processing the images and grasps are
fused into a shared representation, then followed with another 64 neuron hidden
layer before encoding the latent representation. As a regularizer, we inject
white noise at all network inputs with $\sigma=0.05$ and apply weight
normalization \cite{salimans2016weight}.

As mentioned, in order to compute $p_{\theta}(\mathbf{y|x,z})$, we employ a
prediction network for making an initial guess (i.e.
$p_{\theta}(\mathbf{y|x})$), and add it to a prediction of
$p_{\theta}(\mathbf{y|z})$ from the generator network using a sampled
$\mathbf{z}$ (Figure \ref{fig:generator_network}). Our prior network follows a
similar structure to the recognition network; in the CVAE model, we drop the
input $\mathbf{y}$, and only process the visual stream. In the
\textit{recurrent} CVAE (R-CVAE), we take an initial guess made by the
prediction network, and feed it back into the corresponding input in the prior
network.

Two other models were tested. The \textit{Gaussian stochastic neural network}
(GSNN) \cite{sohn2015learning}, which is derived by setting the recognition and
prior network equal
(i.e.~$q_{\theta}(\mathbf{z|x,y})=p_{\theta}(\mathbf{z|x})$), and a baseline
convolutional neural network (CNN) that takes input images and tries to predict
a grasp configuration directly. For evaluating the CLL, we found that 100
samples for importance sampling (R-CVAE and CVAE), and 1000 samples for
Monte-Carlo estimates with the GSNN were sufficient to obtain consistent
estimates.

\section{Results}

\begin{figure}[tb]
\centering
\begin{subfigure}{0.25\textwidth}
  \centering
	\includegraphics[width=0.9\textwidth]{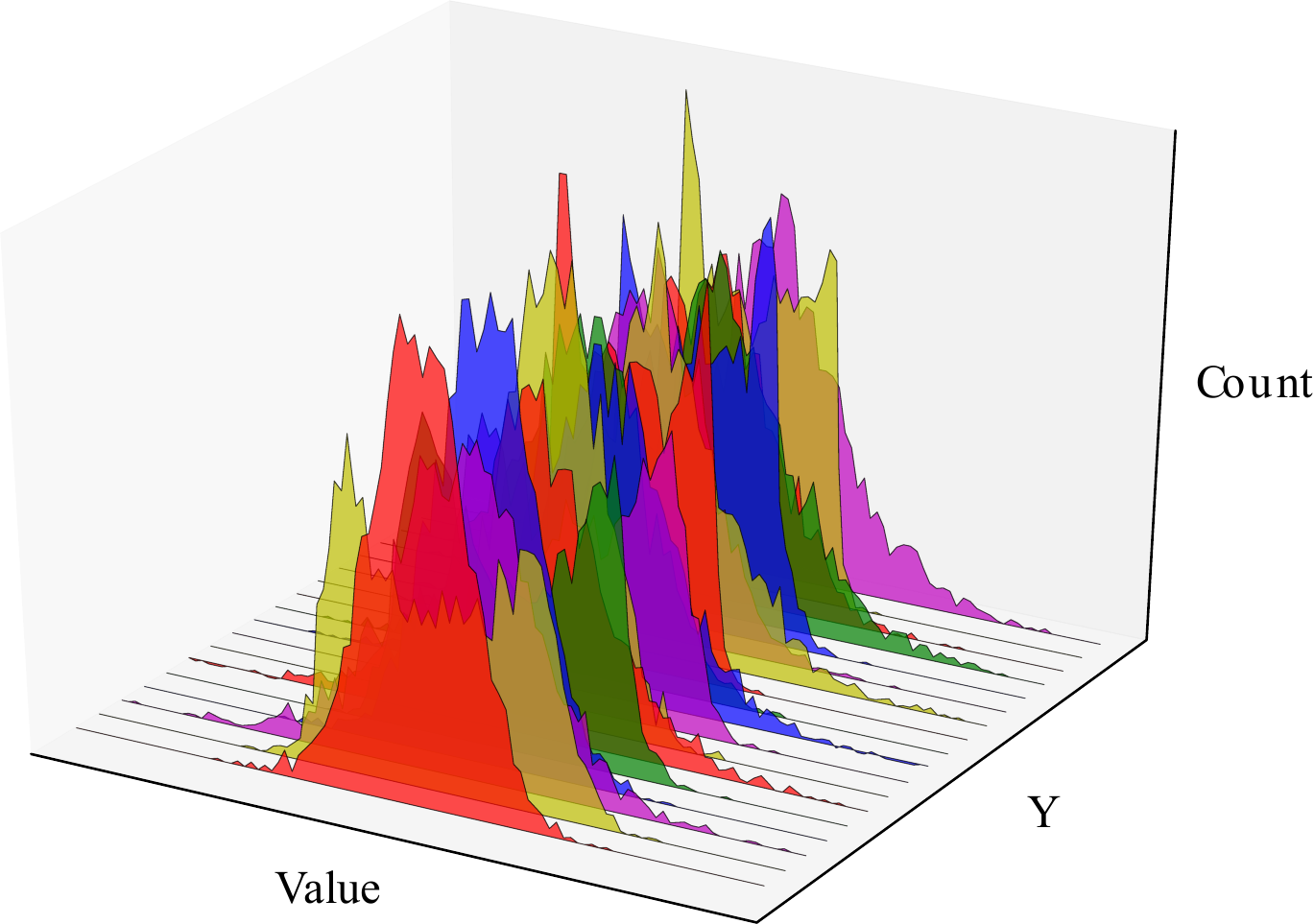}
    \caption{Learned output distributions}
  	\label{fig:output_dist}
\end{subfigure}%
\begin{subfigure}{0.25\textwidth}
  \centering
  \includegraphics[width=0.9\textwidth]{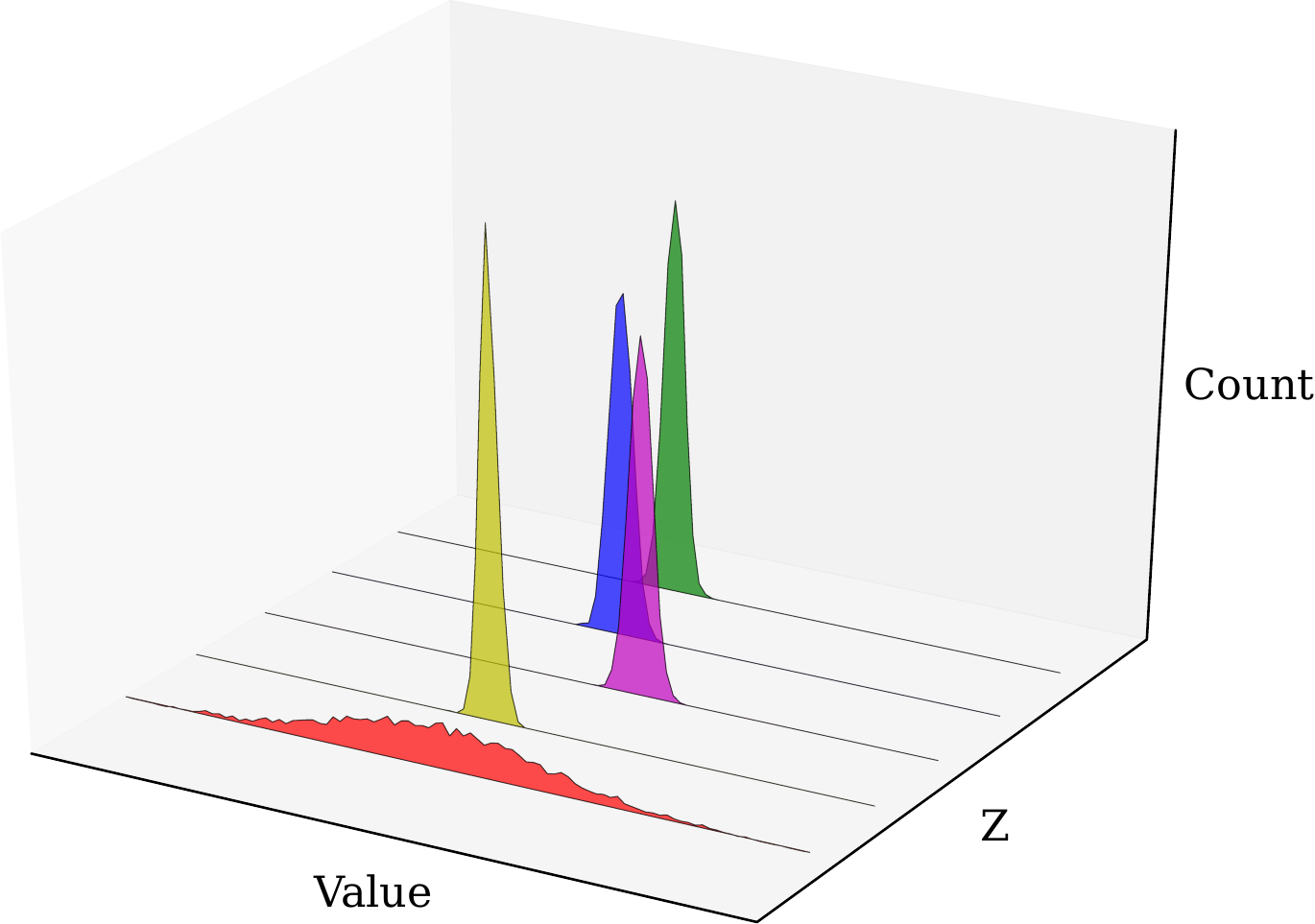}
  	\caption{Learned latent distributions}
    \label{fig:latent_dist}
\end{subfigure}
\caption{Histograms of output (normalized) and latent space samples for a single object-grasp
instance. Histograms generated by drawing 4000 samples from the R-CVAE network.}
\label{fig:learned_distributions}
\end{figure}

To emphasize the fact that the networks are learning \textit{distributions} over
the data instead of deterministic predictions, Figure
\ref{fig:learned_distributions} presents a sample histogram of the latent and
output variable space for a single object-grasp instance.

%
\subsection{Conditional log-likelihood}

Table \ref{tab:negative_cll} presents the estimated negative CLL for each of the
model types, split between the two experimental paradigms: test sets composed of
similar or different objects relative to the training set. The CLL scores for
the CVAE and R-CVAE show that they significantly outperform the GSNN and CNN,
indicating a tighter lower-bound and better approximation to
$\log p(\mathbf{x})$. This result could be due to the parameterization of the
prior distribution; in the CVAE and R-CVAE models, the prior was modulated by
the use of a prior network, allowing the predicted means flexibility in shifting
their inputs. The GSNN, on the other hand, uses a recognition network that only
has information about input visual information $\mathbf{x}$, and is unable to shift
the prior mean based on the grasps $\mathbf{y}$.

\begin{table}[htbp]
  \centering
  \caption{Negative CLL for test sets composed of similar or different objects
    (relative to the training set).}
    \begin{tabular}{l|r|r|r|r|r|r}
    \toprule
          & \multicolumn{3}{c|}{Similar objects (n=4,848)} & \multicolumn{3}{c}{Different objects  (n=8,851) } \\
    \midrule
    Train size & 16,384 & 32,768 & 42,351 & 16,384 & 32,768 & 42,351 \\
    \midrule
    CNN   & \multicolumn{1}{r}{24.721} & \multicolumn{1}{r}{24.833} & 24.577 & \multicolumn{1}{r}{26.910} & \multicolumn{1}{r}{26.920} & 26.599 \\
    GSNN  & \multicolumn{1}{r}{22.827} & \multicolumn{1}{r}{22.292} & 21.831 & \multicolumn{1}{r}{27.945} & \multicolumn{1}{r}{32.513} & 33.461 \\
    CVAE  & \multicolumn{1}{r}{15.325} & \multicolumn{1}{r}{13.531} & 13.216 & \multicolumn{1}{r}{18.356} & \multicolumn{1}{r}{18.808} & 16.525 \\
    R-CVAE & \multicolumn{1}{r}{\textbf{13.670}} & \multicolumn{1}{r}{\textbf{13.024}} & \textbf{12.511} & \multicolumn{1}{r}{\textbf{14.277}} & \multicolumn{1}{r}{\textbf{14.128}} & \textbf{13.514} \\
    \bottomrule
    \end{tabular}%
  \label{tab:negative_cll}%
\end{table}%
%
%
\subsection{Simulator}

To evaluate how the network predictions transfer back to the ``picking'' task,
we evaluate a single prediction (again using the distribution means for the
stochastic networks) in the simulator. Given that the task is to pick up an
object, we define a grasp to be \emph{successful} if the object is still held
within the gripper at the height of the picking operation. If the object is no
longer held by the fingertips, contacting other components, or the gripper
failed to contact the object during initial finger placement, the grasp
is deemed a failure.

In order to position the manipulator for each initial grasp, our grasp planner
consists of calculating an optimal initial wrist placement by minimizing the
distance of each of the manipulator's fingers to the predicted target positions:
\begin{equation}
    \label{eqn:minimize_pos}
    \min_{\alpha,\beta,\gamma,T_x,T_y,T_z} \sum_{i=1}^{N} (C_i-Y_i)^2
\end{equation}
\noindent where $\alpha, \beta, \gamma$ are the yaw, pitch, and roll rotational
components, while $T_x$, $T_y$, $T_z$ are the $x$, $y$, and $z$ translational
components. In this optimization, $N$ is the number of fingertips, and $C_{i},
Y_{i}$ are the ground-truth and predicted fingertip positions relative to a
common frame. The results for similar and different objects can be seen in
Table \ref{tab:postgrasp}.

\begin{table}[htbp]
  \centering
  \caption{Percentage of successful simulated grasps.}
    \begin{tabular}{l|r|r|r|r|r|r}
    \toprule
          & \multicolumn{3}{c|}{Similar objects (n=4,848)} & \multicolumn{3}{c}{Different objects  (n=8,851) } \\
    \midrule
    Train size & 16,384 & 32,768 & 42,351 & 16,384 & 32,768 & 42,351 \\
    \midrule
    CNN   & \multicolumn{1}{r}{0.155} & \multicolumn{1}{r}{0.202} & 0.199 & \multicolumn{1}{r}{0.106} & \multicolumn{1}{r}{0.138} & 0.145 \\
    GSNN  & \multicolumn{1}{r}{0.169} & \multicolumn{1}{r}{0.177} & 0.190 & \multicolumn{1}{r}{0.115} & \multicolumn{1}{r}{0.145} & 0.147 \\
    CVAE  & \multicolumn{1}{r}{\textbf{0.344}} & \multicolumn{1}{r}{\textbf{0.346}} & 0.347 & \multicolumn{1}{r}{\textbf{0.302}} & \multicolumn{1}{r}{0.301} & 0.295 \\
    R-CVAE & \multicolumn{1}{r}{0.318} & \multicolumn{1}{r}{0.323} & \textbf{0.362} & \multicolumn{1}{r}{0.288} & \multicolumn{1}{r}{\textbf{0.304}} & \textbf{0.315} \\
    \bottomrule
    \end{tabular}%
  \label{tab:postgrasp}%
\end{table}%

Two key results can be seen from this table. First, grasp predictions
made by the baseline CNN appear to be unable to match that of the generative
CVAE and R-CVAE models. It is likely that this result stems from learning a
discriminative model with a multimodal setting. Partial success for the CNN's
predictions may also be due in part to a grasp planner that succeeds under
fairly weak predictions.

Second, with respect to data efficiency, the relative gains for the baseline CNN
model (between the 16k and 42k training set sizes) appears to be much greater
then the generative CVAE and R-CVAE models. The literature has reported advances
in supervised learning that have been able to leverage very large labeled
training datasets, but the gains for unsupervised learning methods are less
documented.

\subsection{Multimodal grasp distributions}

\begin{figure*}[!htb]
\centering
\includegraphics[width=0.85\textwidth]{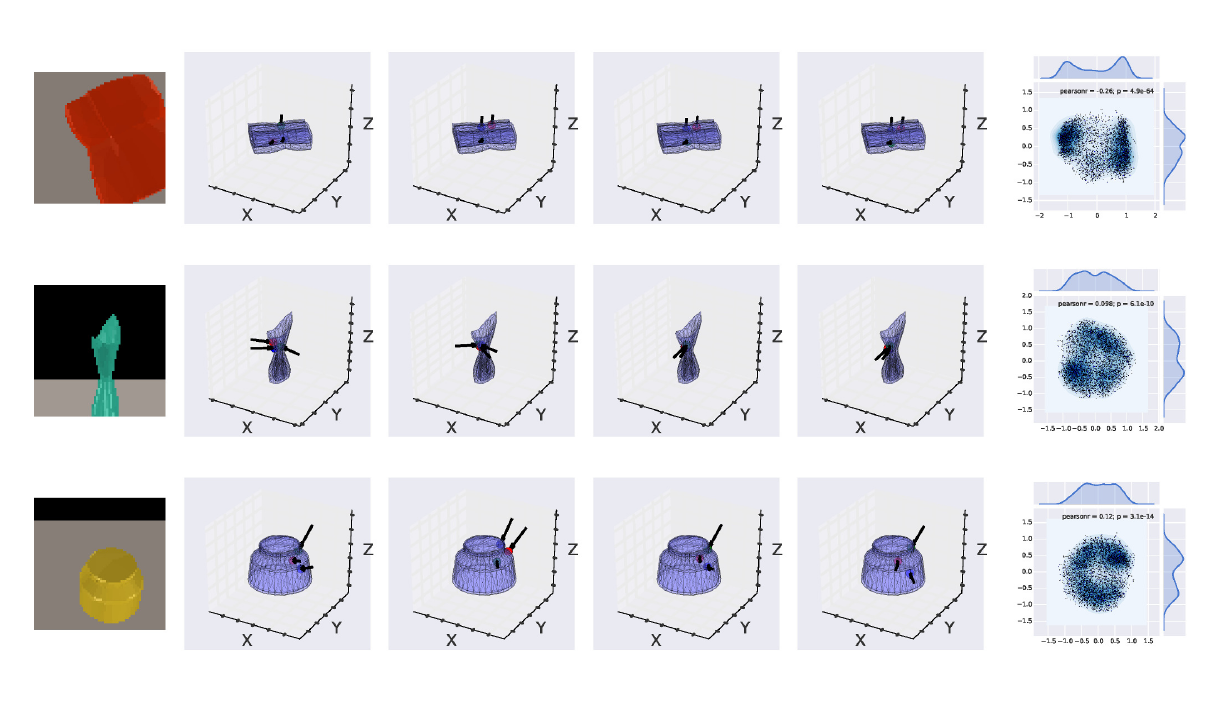}
\caption{Sampled grasps from similar objects. \textbf{Left}: RGB image of
            object, \textbf{Middle}: Plotted grasp configurations (positions and normals),
            \textbf{Right}: t-SNE plot of learned grasp space with superimposed
            kernel density estimate (note multiple discrete modes).\looseness=-1}
\label{fig:sampled_grasps_sim}
\includegraphics[width=0.85\textwidth]{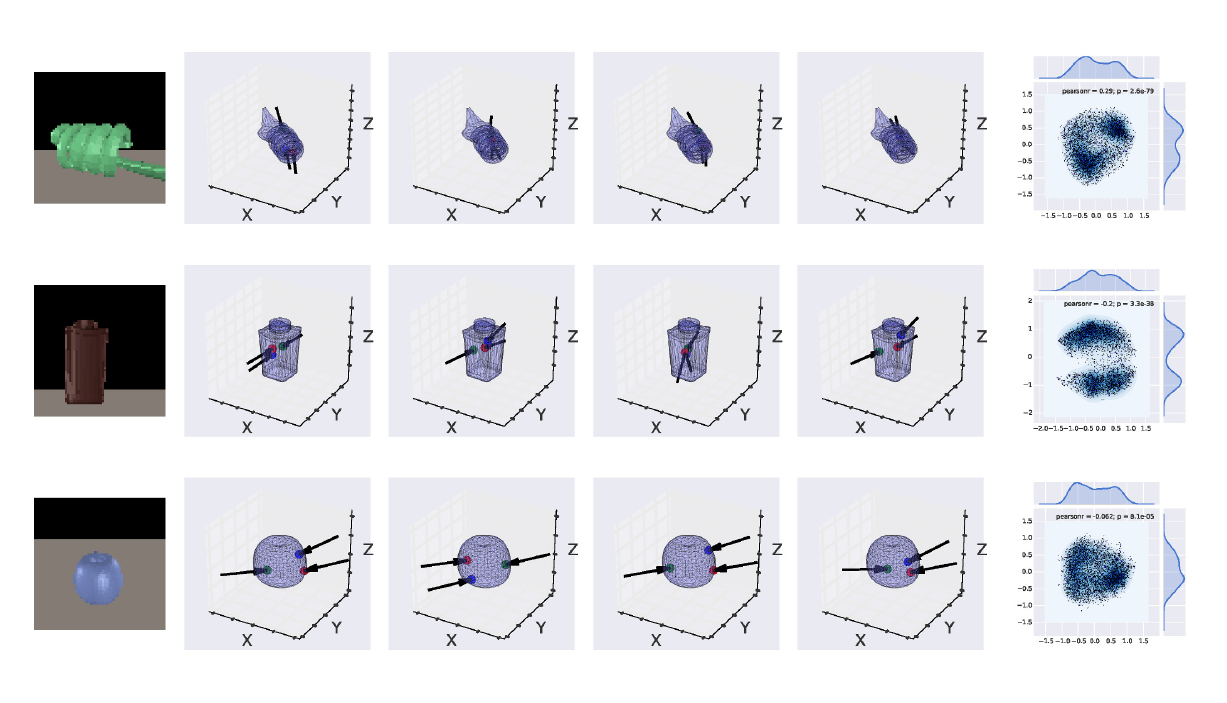}
\caption{Sampled grasps from different objects. \textbf{Left}: RGB image of
             object, \textbf{Middle}: Plotted grasp configurations (positions and normals),
             \textbf{Right}: t-SNE plot of learned grasp space with superimposed
             kernel density estimate (note multiple discrete modes).\looseness=-1}
\label{fig:sampled_grasps_diff}
\end{figure*}

In Figures \ref{fig:sampled_grasps_sim} \& \ref{fig:sampled_grasps_diff}, we
demonstrate the learned multimodality of our networks by sampling different
grasps from a single grasp instance\footnote{The plotted grasp configuration is
the result of solving for that grasp that maximizes: $\mathbf{y^*}=\argmax_y
\frac{1}{L} \sum_{l=1}^{L} p_{\theta}(\mathbf{y|x,z}^{(l)})$, using
$\mathbf{z}^{(l)} \sim p_{\theta}(\mathbf{z|x})$ and $L$ = 50.}.
In these figures, we present the input RGB image (left), the plotted grasp
configuration \& object (middle), as well as a t-SNE
\cite{maaten2008visualizing} plot of the learned grasp space (right). t-SNE is a
method for visualizing high-dimensional data by projecting it down to a
lower-dimensional space. In these plots, one can clearly see distributions with
multiple modes, which in turn appear to be reflected in the plotted grasp space.

\section{Discussion}

Further inspection of the grasp space in Figures \ref{fig:sampled_grasps_sim} \&
\ref{fig:sampled_grasps_diff} appears to show that many of the grasps share a
similar vertical distribution. We believe this may be a result of the data
collection process, where reachability constraints prevent certain grasps from
being executed (e.g.~on the lower part of objects due to contact with the
table).

We have identified a few potential limitations of our current approach. First,
our method still requires tens of thousands of examples to train, which is
expensive to collect in the real world. Second, our evaluation strategy has only
focused on objects with fixed intrinsic properties, which is a simplification of
real-world characteristics. Compared to feed-forward networks, a sampling-based
approach is more computationally expensive and there may be alternate ways of
simplifying its computational requirements. Finally, there are also other practical considerations that
could be taken during data-collection, such as optimizing the closing strategy
of the manipulator for e.g., more reactive grasping.

\section{Conclusion}

In this work we presented a conceptual framework for robotic grasping, the
\textit{grasp motor image}, which integrates perceptual information and grasp
configurations using deep generative models. Applying our method to a simulated
grasping task, we demonstrated the capacity of these models to transfer learned
knowledge to novel objects under varying amounts of available training data, as
well as their strength in capturing multimodal data distributions.

Our primary interest moving forward is in investigating how objects with
variable intrinsic properties can be learned with the GMI, specifically, by
introducing additional sensory modalities into the system. We are also
interested in investigating how continuous contact with an object contributes to
the formation of the GMI. We find work within the cognitive sciences on the
effects of somatosensory input on motor imagery
(c.f. \cite{mizuguchi2015effect}) to be an interesting starting point.

\bibliographystyle{IEEEtran}
\bibliography{IEEEabrv,bib}

\end{document}